\documentclass[runningheads]{llncs}
\usepackage[T1]{fontenc}
\usepackage{graphicx,verbatim}
\usepackage{booktabs}
\usepackage{multirow}
\usepackage{amsmath}
\usepackage{amssymb}
\usepackage[table,xcdraw]{xcolor}

\begin{document}
\title{Improving Factuality of 3D Brain MRI Report Generation with Paired Image-domain Retrieval and Text-domain Augmentation}

\titlerunning{Factual 3D MRI Report Generation via PIRTA}
%

\author{
Junhyeok Lee\inst{1}$^{\#}$\thanks{$^{\#}$\,Co-first authors. \quad $^{\dagger}$\,Co-corresponding authors.} \and
Yujin Oh\inst{2,3}$^{\#}$ \and
Dahyoun Lee\inst{3} \and
Hyon Keun Joh\inst{4} \and
Minchul Kim\inst{5} \and
Chul-Ho Sohn\inst{6,7} \and
Sung Hyun Baik\inst{6,8} \and
Cheolkyu Jung\inst{6,8} \and
Jung Hyun Park\inst{6,9} \and
Kyu Sung Choi\inst{6,7}$^{\dagger}$ \and
Byung-Hoon Kim\inst{3,10,11,12}$^{\dagger}$ \and
Jong Chul Ye\inst{13}$^{\dagger}$
}
\authorrunning{J. Lee, Y. Oh et al.}
\institute{
Cancer Biology, Seoul National University College of Medicine, Korea \and
Radiology, Massachusetts General Hospital and Harvard Medical School, USA \and
Biomedical Systems Informatics, Yonsei University College of Medicine, Korea \and
Graduate School, Yonsei University, Korea \and
Radiology, Kangbuk Samsung Hospital, Korea \and
Radiology, Seoul National University College of Medicine, Korea \and
Radiology, Seoul National University Hospital, Korea \and
Radiology, Seoul National University Bundang Hospital, Korea \and
Radiology, SMG-SNU Boramae Medical Center, Korea \and
Psychiatry, Yonsei University College of Medicine, Korea \and
Behavioral Sciences in Medicine, Yonsei University College of Medicine, Korea \and
Yonsei Institute for Digital Health, Korea \and
Kim Jaechul Graduate School of AI, KAIST, Korea\\
\email{ent1127@snu.ac.kr; egyptdj@yonsei.ac.kr; jong.ye@kaist.ac.kr}
}
\maketitle

\begin{abstract}
Acute ischemic stroke (AIS) requires time-critical decision-making, where inaccurate interpretation of neuroimaging findings can lead to irreversible disability. 
Diffusion-weighted imaging (DWI) and apparent diffusion coefficient (ADC) maps from magnetic resonance imaging (MRI) are central to detecting acute infarction, yet generating factually reliable radiology reports directly from 3D MRI remains challenging due to the difficulty of learning robust cross-modal alignments between volumetric images and clinical text.
We propose paired image-domain retrieval and text-domain augmentation (PIRTA), a retrieval-augmented generation framework that improves report factuality by avoiding explicit image--text alignment. 
PIRTA retrieves clinically similar 3D DWI/ADC volumes using a pretrained 3D vision encoder and leverages their paired clinician-authored reports to ground large language model (LLM)–based report generation.
Experiments on multi-institutional in-house data, a held-out external privacy-preserving cohort, and the public ISLES benchmark demonstrate that PIRTA achieves strong image-domain retrieval performance and consistently improves ischemic-territory accuracy, a clinically grounded surrogate for report factuality, compared to direct image-to-text baselines. 
These results indicate that retrieval-grounded generation provides a scalable and reliable paradigm for producing factually consistent radiology reports from complex 3D brain MRI.
Source code is available at \url{https://github.com/jhlee0619/PIRTA}.

\end{abstract}

\section{Introduction}
\label{sec:introduction}

Stroke is the second leading cause of death worldwide and a major contributor to long-term disability and dementia~\cite{feigin2022world}. Acute ischemic stroke (AIS), which accounts for approximately 70\% of all stroke cases, results from abrupt interruption of cerebral blood flow and requires time-critical intervention, as treatment delays substantially increase the risk of irreversible neurological deficits~\cite{campbell2019ischaemic}. Accordingly, current American Heart Association/American Stroke Association (AHA/ASA) guidelines emphasize rapid reperfusion therapy within narrow therapeutic windows to improve functional outcomes~\cite{powers2019guidelines}.

Magnetic resonance imaging (MRI) plays a central role in AIS diagnosis and management by delineating tissue viability and ischemic injury~\cite{san2018imaging}. In particular, diffusion-weighted imaging (DWI) and apparent diffusion coefficient (ADC) maps enable sensitive detection of acute infarction and support assessment of hemorrhagic transformation risk~\cite{campbell2019ischaemic}. These findings are summarized in radiology reports, which provide the clinically actionable interpretation guiding thrombolysis or thrombectomy decisions. While recent advances in artificial intelligence (AI) have enabled automated AIS detection and outcome prediction from MRI~\cite{mouridsen2020artificial}, most prior work has focused on classification or segmentation tasks rather than generating clinically usable radiology reports. Existing report generation approaches typically rely on supervised learning with extensive annotation or segmentation-based pipelines, limiting scalability and cross-institutional generalization~\cite{herzog2020integrating,cetinoglu2021detection,tasci2022deep,lee2023automatic,koska2024deep,Liu2023-ej}.

\begin{figure}[t]
\centering
\includegraphics[width=0.8\linewidth]{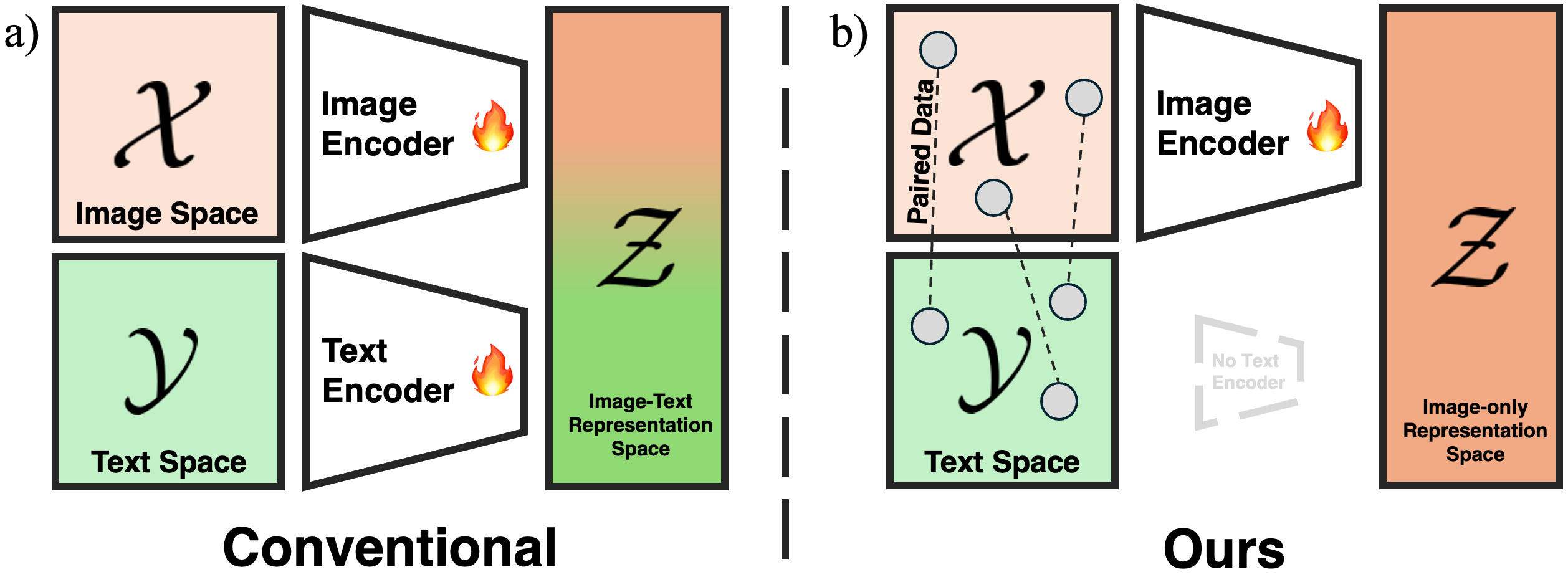}
\caption{Conceptual comparison of cross-modal alignment strategies. (a) Conventional methods align image and text encoders in a shared representation space, requiring joint modeling across modalities. (b) PIRTA avoids explicit cross-modal alignment by retrieving similar images in the image domain, and using their paired clinician-authored reports for text-domain augmentation.}
\label{fig:intro}
\end{figure}

Radiology report generation from MRI is inherently challenging because it requires translating high-dimensional volumetric image evidence into structured clinical language. Although vision--language models (VLMs) and large language models (LLMs) have demonstrated strong performance on 2D modalities such as chest X-ray and CT~\cite{singhal2023towards,lee2023llm,li2023llava}, reliable report generation from 3D MRI remains limited by the scarcity of pretrained 3D medical image encoders and the difficulty of learning robust cross-modal alignment between volumetric image representations and radiology text (Fig.~\ref{fig:intro}a)~\cite{Huo_2024_CVPR,ma2024learningmodalityknowledgealignment}.

To mitigate these challenges, we propose paired image-domain retrieval and text-domain augmentation (PIRTA), a retrieval-augmented generation (RAG) framework~\cite{lewis2020retrieval} that improves report factuality by avoiding explicit image--text alignment (Fig.~\ref{fig:intro}b). PIRTA retrieves clinically similar 3D DWI/ADC volumes from an in-domain paired image--report database using a pretrained 3D vision encoder, and leverages their paired clinician-authored reports as evidence to guide LLM-based report generation. To enable robust 3D retrieval, we pretrain a 3D Vision Transformer (ViT)~\cite{dosovitskiy2020image} with MAE-based self-supervised learning~\cite{he2022masked} on large-scale unlabeled MRI.

Our contributions are as follows:
\begin{itemize}
    \item We propose PIRTA, an image-domain retrieval and text-domain augmentation framework that avoids explicit cross-modal alignment, reducing optimization complexity for 3D MRI report generation.
    \item We develop a robust 3D DWI/ADC encoder through large-scale MAE pretraining, demonstrating significant improvements in cross-institution retrieval generalization.
    \item We evaluate factuality using clinically actionable ischemic-territory accuracy on multi-institutional internal data and public benchmarks, showing that PIRTA improves factual consistency by grounding generation in retrieved clinical evidence.
\end{itemize}

\section{Methods}

\subsection{Problem Definition and Cross-modal Learning Complexity}

Let $\mathcal{X}$ denote the space of 3D brain MRI volumes and $\mathcal{Y}$ the space of radiology reports; given a query image $x_q\in\mathcal{X}$, the objective is to generate or retrieve a clinically accurate report $y\in\mathcal{Y}$. Conventional image-to-text approaches jointly train an image encoder $f_{\text{image}}:\mathcal{X}\rightarrow\mathcal{Z}$ and a text encoder $f_{\text{text}}:\mathcal{Y}\rightarrow\mathcal{Z}$ into a shared latent space $\mathcal{Z}$ by minimizing a distance between paired embeddings,
\begin{equation}
\min d\big(f_{\text{image}}(x_q),f_{\text{text}}(y)\big),
\end{equation}
implicitly learning the joint distribution $P(x,y)$ over a joint hypothesis space $\mathcal{H}=\mathcal{H}_{\text{image}}\times\mathcal{H}_{\text{text}}$ with complexity $C(\mathcal{H})=C(\mathcal{H}_{\text{image}})+C(\mathcal{H}_{\text{text}})$~\cite{shalev2014understanding}. This is particularly challenging for 3D medical imaging due to limited paired data and the representational gap between volumetric MRI and clinical text.

\subsection{PIRTA: Paired Image-domain Retrieval and Text-domain Augmentation}

\begin{figure*}[t]
\centering
\includegraphics[width=\linewidth]{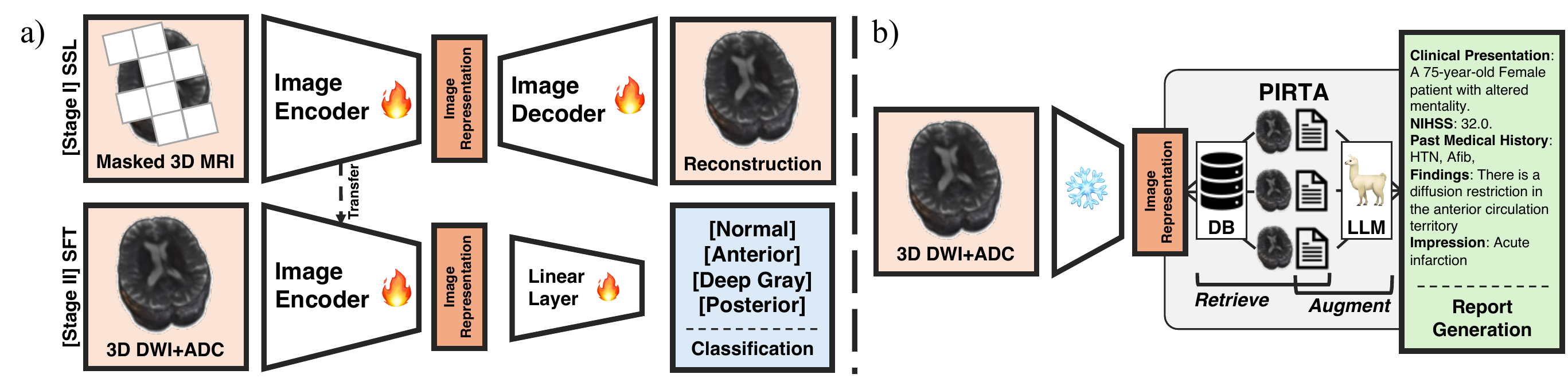}
\caption{Overview of PIRTA. (a) Training the 3D MRI image encoder: MAE-based SSL pretraining on unlabeled data followed by supervised fine-tuning for ischemic-territory classification. (b) Cross-modal RAG with PIRTA: using the frozen encoder to retrieve similar images and their paired reports, which augment LLM-based report generation.}
\label{fig:scheme}
\end{figure*}

PIRTA reformulates report generation as in-domain image retrieval followed by text-domain augmentation. Given a paired database $\mathcal{D}=\{(x_i,y_i)\}_{i=1}^m$ of MRI volumes and clinician-authored reports, we train only an image encoder $f_{\text{image}}:\mathcal{X}\rightarrow\mathcal{Z}$. For a query $x_q$, PIRTA retrieves the most relevant case by minimizing cosine distance in the embedding space,
\begin{equation}
x^*=\arg\min_{x_i\in\mathcal{D}} d\big(f_{\text{image}}(x_q),f_{\text{image}}(x_i)\big),
\end{equation}
and uses the paired report $y^*$ as external evidence for report generation. This reduces the hypothesis space to $\mathcal{H}_{\text{PIRTA}}=\mathcal{H}_{\text{image}}$, focusing learning on the marginal $P(x)$ rather than the joint $P(x,y)$. Grounding generation in clinician-authored reports retrieved purely in the image space anchors outputs in expert narratives and mitigates ungrounded generation, while reducing data and optimization burden. Fig.~\ref{fig:scheme} overviews PIRTA, including MAE-based pretraining and retrieval-grounded report generation.

\subsection{3D MRI Image Encoder Training}

The image encoder is trained in two stages: self-supervised pretraining followed by supervised fine-tuning. For pretraining, we adopt a masked autoencoder (MAE) framework~\cite{he2022masked} with a 3D Vision Transformer (ViT)~\cite{dosovitskiy2020image} backbone. DWI and ADC volumes are concatenated as a two-channel 3D input and partitioned into non-overlapping volumetric patches, a subset of which is randomly masked. The encoder processes the visible patches, while a lightweight decoder reconstructs the masked patches using mean squared error loss.

Following pretraining, the encoder is fine-tuned for ischemic-territory classification (anterior, deep gray, posterior, normal). The MAE decoder is replaced by a linear classification head, and the model is optimized using cross-entropy loss on labeled internal datasets. This two-stage training yields a 768-dimensional volumetric embedding for retrieval.

\subsection{Retrieval-based Report Generation}

After fine-tuning, latent embeddings of all training images are extracted to construct the retrieval database $\mathcal{D}$. Given a query MRI volume, cosine similarity is computed between its embedding and all database embeddings, and the top-$k$ most similar cases are retrieved. The paired radiology reports associated with the retrieved images are then used to augment report generation by a large language model (LLM).

Specifically, we fine-tune a large language model using low-rank adaptation (LoRA)~\cite{hu2021lora} on structured radiological findings. During inference, retrieved reports and their similarity scores are injected into the prompt as external evidence, guiding the LLM to generate a final report. The report is structured into Findings and Impression sections and includes metadata fields (clinical presentation, NIHSS, and PMH) when available.

\section{Experiments}
\subsection{Datasets}
\label{sec:data}
We use multi-institutional in-house MRI data from 1,831 adults (993 acute ischemic stroke cases and 838 controls) for model training and internal evaluation (Internal Cohort). 
External generalization is assessed using two independent datasets acquired at centers not included in the Internal Cohort: a held-out, privacy-preserving dataset (External Cohort 1, $n=580$) and the public multi-site ISLES benchmark (External Cohort 2, $n=206$)~\cite{hernandez2022isles}. 
For self-supervised pretraining, we leverage unlabeled DWI/ADC volumes from 38,532 UK Biobank subjects~\cite{alfaro2018image}.
All reports were transformed into structured findings templates curated by a certified clinical expert.
Train/validation/test partitions are enforced strictly at the patient level, with no patient appearing in multiple splits; the retrieval database is built from training-set embeddings only, so validation and test images are never used as retrieval candidates.

\subsection{Implementation details}
\label{sec:imple}
The image encoder is pretrained with MAE-based SSL and fine-tuned for four-class territory classification on internal data. Retrieval uses cosine similarity across training-set embeddings to identify top-$k=5$ neighbors. Report generation is performed using LLaMA3-8B-Instruct~\cite{dubey2024llama}, instruction-tuned with LoRA on structured findings. LLM fine-tuning is conducted for 5 epochs with a batch size of 8 and LoRA rank $r=16$. For inference, retrieved findings and their similarity scores are provided as textual evidence in the LLM prompt.

\begin{table}[t]
\centering
\fontsize{7pt}{8pt}\selectfont
\caption{Image retrieval performance (\%) versus pretraining size. \textit{None}: random initialization; \textit{Small}: MAE on the internal cohort only ($n{=}1{,}831$); \textit{Large}: MAE on $n{=}38{,}532$ UK Biobank subjects.}
\label{tab:comp_mri}
\begin{tabular}{llcccccc}
\toprule
Dataset & Pretrain & mAP@1 & mAP@5 & mAP@10 & Acc@1 & Acc@5 & Acc@10 \\ 
\midrule
\multirow{3}{*}{Internal Cohort}
& None  & 80.73 & 81.90 & 81.57 & 80.73 & 84.40 & 84.40 \\
& Small & 90.37 & 90.78 & 90.85 & 90.37 & 91.74 & 92.66 \\
& Large & \textbf{94.04} & \textbf{94.27} & \textbf{94.15} & \textbf{94.04} & \textbf{94.50} & \textbf{94.50} \\ 
\midrule
\multirow{3}{*}{External Cohort 1}
& None  & 56.72 & 58.22 & 57.94 & 56.72 & 61.55 & 64.48 \\
& Small & 68.97 & 69.19 & 69.11 & 68.97 & 70.17 & 70.69 \\
& Large & \textbf{71.21} & \textbf{71.43} & \textbf{71.30} & \textbf{71.21} & \textbf{71.72} & \textbf{71.72} \\ 
\midrule
\multirow{3}{*}{External Cohort 2}
& None  & 38.83 & 40.48 & 40.42 & 38.83 & 44.17 & 48.06 \\
& Small & 65.53 & 66.26 & 66.39 & 65.53 & 67.48 & 69.42 \\
& Large & \textbf{70.87} & \textbf{71.52} & \textbf{71.24} & \textbf{70.87} & \textbf{72.33} & \textbf{72.82} \\
\bottomrule
\end{tabular}
\end{table}

\begin{figure*}[t]
\centering
\includegraphics[width=\linewidth, trim=0 11.6cm 0 0, clip]{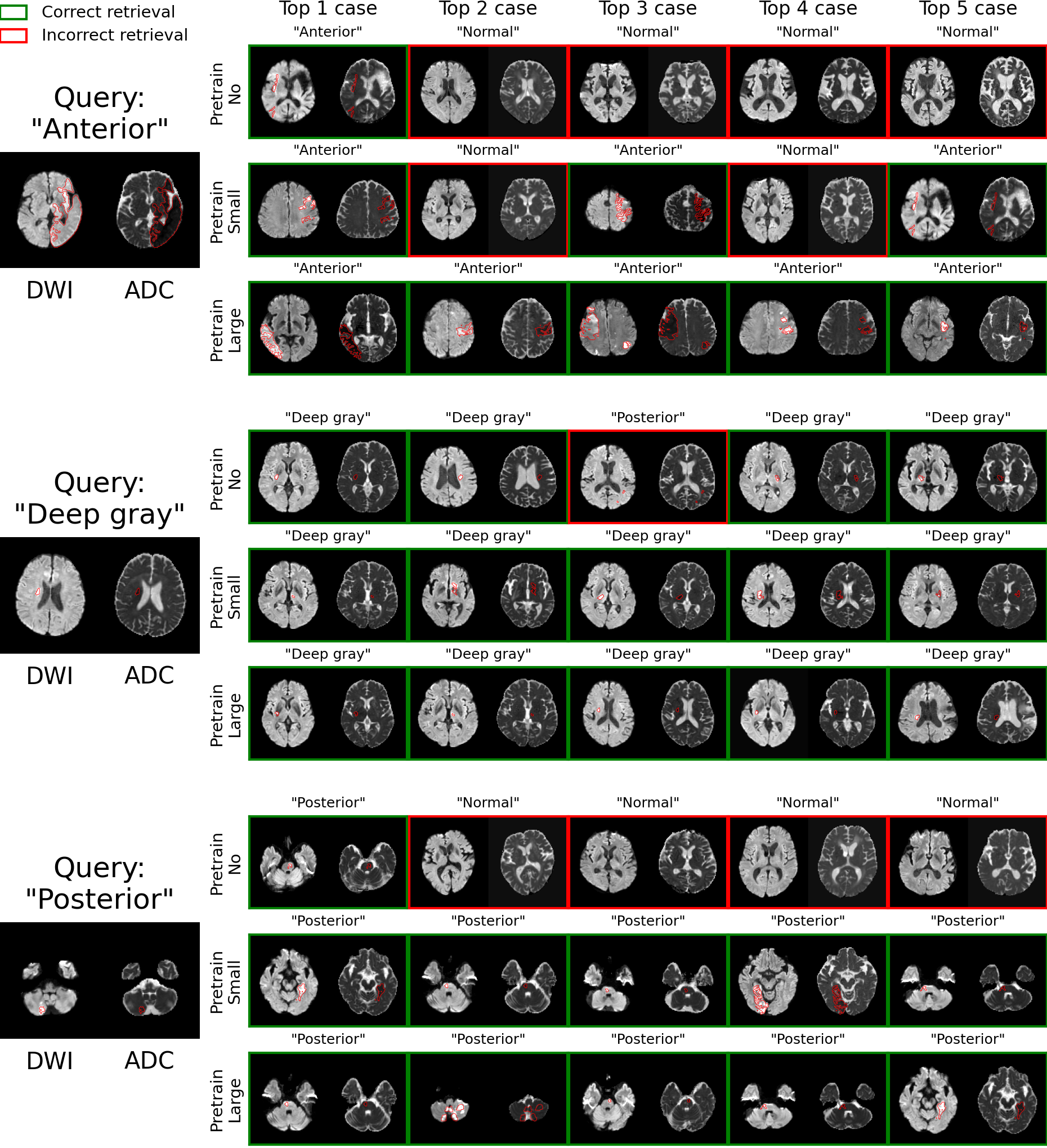}
\caption{Qualitative retrieval examples for a query volume (left) and the top-5 retrieved scans (right). Green indicates correct retrieval; red indicates incorrect retrieval.}
\label{fig:retrieval_example}
\end{figure*}

\section{Results}
\subsection{Image-domain retrieval and representation generalization}

We evaluate image-domain retrieval performance using mean average precision (mAP@$k$) and top-$k$ accuracy (Acc@$k$), as summarized in Table~\ref{tab:comp_mri}.
While strong retrieval performance is achieved on the Internal Cohort (mAP@1 of 94.0\%), the most pronounced gains are observed on the external cohorts, particularly External Cohort 2 (ISLES), indicating improved robustness under substantial domain shift.

To further assess representation quality, we conduct an auxiliary sanity check using ischemic-territory classification with the same encoder (Table~\ref{tab:class_mri}). 
Consistent with the retrieval results, large-scale MAE pretraining substantially improves multi-class classification accuracy across both internal and external cohorts, with especially large improvements on External Cohort 2. 
These parallel trends in retrieval and classification performance suggest that the encoder learns semantically meaningful and transferable 3D representations, rather than task-specific heuristics.

\begin{table}[t!]
\caption{Effect of MAE pretraining scale on multi-class territory classification accuracy (Acc@1), shown as a sanity check of representation quality. \textit{None}: random initialization; \textit{Small}: MAE on the internal cohort ($n{=}1{,}831$); \textit{Large}: MAE on $n{=}38{,}532$ UK Biobank subjects.}
\centering
\fontsize{7pt}{8pt}\selectfont
\begin{tabular}{llccccc}
\toprule
\multirow{2}{*}{Dataset} & \multirow{2}{*}{Pretrain} & \multicolumn{4}{c}{Class-wise Acc@1} & \multirow{2}{*}{\begin{tabular}[c]{@{}c@{}}Multi-class\\ Acc@1\end{tabular}} \\ \cmidrule(lr){3-6}
& & Normal & Anterior & Deep gray & Posterior & \\ 
\midrule
\multirow{3}{*}{Internal Cohort} & None & 96.3 & 87.96 & 89.81 & 89.81 & 81.94 \\
& Small & 99.54 & 93.06 & 91.67 & 95.37 & 89.81 \\
& Large & \textbf{100.0} & \textbf{95.37} & \textbf{93.98} & \textbf{96.76} & \textbf{93.06} \\ 
\midrule
\multirow{3}{*}{External Cohort 1} & None & 60.87 & 78.26 & 86.29 & 83.95 & 53.18 \\
& Small & 69.23 & 89.8 & 90.47 & 85.95 & 66.22 \\
& Large & \textbf{72.24} & \textbf{90.64} & \textbf{90.8} & \textbf{87.79} & \textbf{69.23} \\ 
\midrule
\multirow{3}{*}{External Cohort 2} & None & - & 69.52 & 76.67 & 68.1 & 40.0 \\
& Small & - & \textbf{84.29} & \textbf{86.67} & 80.95 & 61.9 \\
& Large & - & 82.38 & 84.76 & \textbf{83.81} & \textbf{68.1} \\
\bottomrule
\end{tabular}
\label{tab:class_mri}
\end{table}

\begin{figure}[t]
\centering
\includegraphics[width=0.95\linewidth]{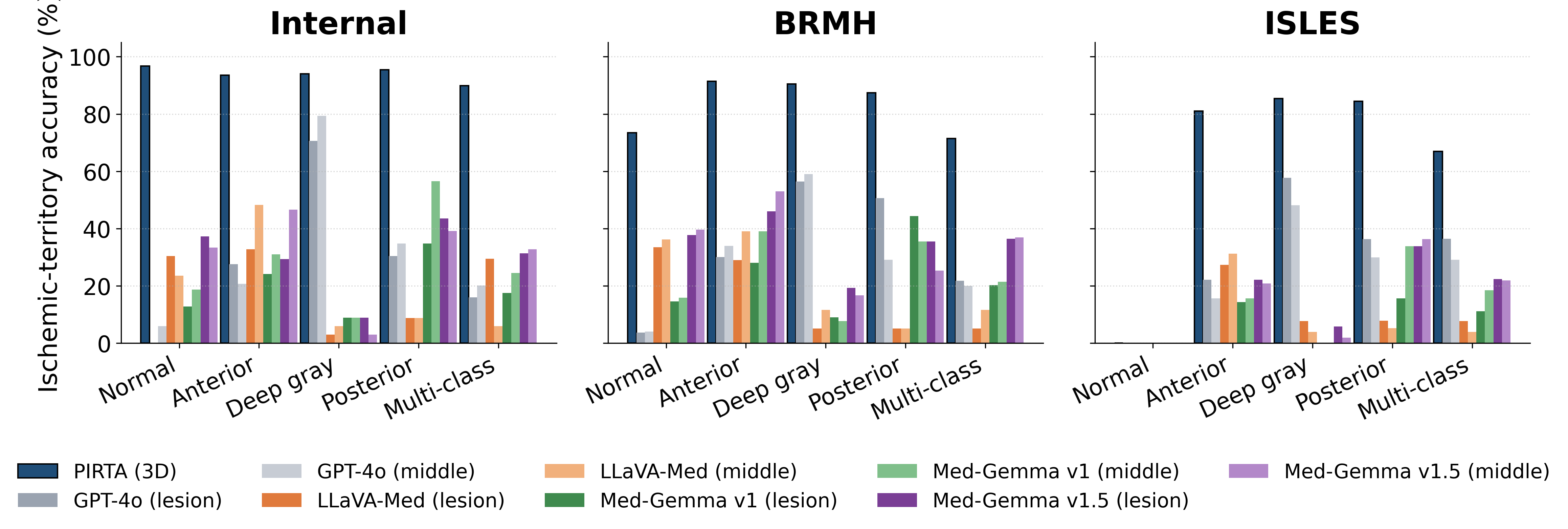}
\caption{Ischemic-territory accuracy for report generation. PIRTA (3D) is compared with 2D-slice baselines (GPT-4o, LLaVA-Med, and the medical-domain VLM Med-Gemma v1/v1.5). \textit{Multi-class} denotes patients with $\geq$2 affected territories.}
\label{fig:performance_graph}
\end{figure}

\subsection{Clinical factuality evaluation of generated reports}
\label{sec:result_llm}

We evaluate report factuality using ischemic-territory accuracy, a clinically actionable attribute that directly influences treatment decisions in acute stroke care.

We compare PIRTA with both general-purpose 2D multimodal baselines (GPT-4o, LLaVA-Med~\cite{li2023llava}) and a medical-domain 2D vision--language model (Med-Gemma v1 and v1.5~\cite{medgemma2024}), each receiving an axial slice (mid-brain or lesion-centered when available) and the identical output-format prompt.
PIRTA achieves substantially higher ischemic-territory accuracy than every 2D baseline (Fig.~\ref{fig:performance_graph}), leading the strongest 2D baseline by +57.2 / +34.5 / +30.6 multi-class accuracy points on the internal cohort, BRMH, and ISLES, with per-class gaps persisting across territories.
This indicates that retrieval-grounded 3D evidence, not merely medical-domain pretraining of a 2D backbone, is the dominant factuality contributor, with failures concentrated at low-similarity, anatomically rare infarct patterns.

\begin{table*}[t!]
\centering
\fontsize{8pt}{9.5pt}\selectfont
\caption{
Qualitative assessment of text-domain augmentation. 
Correct ischemic territory information and corresponding similarity scores are highlighted in \textcolor{blue}{blue}
}
\label{tab:qualitative}
\begin{tabular}{p{0.17\linewidth} | p{0.77\linewidth}}
\toprule[1.5pt]
\begin{center}
\vspace*{-1.5em}
\textbf{DWI \& ADC} \\
\vspace{0.7em}
\includegraphics[width=0.16\textwidth]{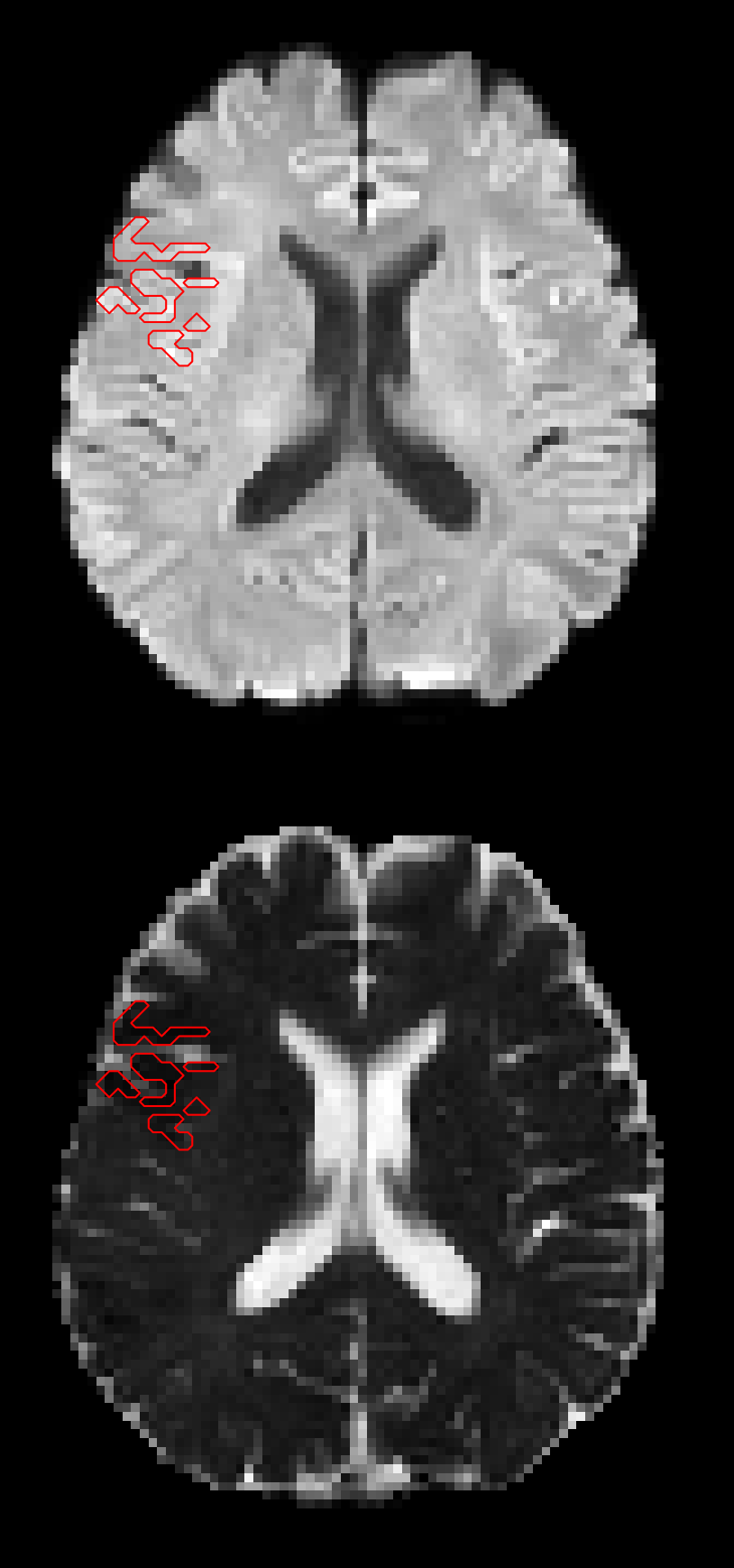}
\vspace*{-1.5em}
\end{center}
&
\begin{minipage}[t]{\linewidth}
\fontsize{7pt}{8pt}\selectfont
\vspace*{-0.5em}
\textbf{Ground-truth:} \\
\textbf{Clinical Presentation:} A 58-year-old male patient. \\
\textbf{NIHSS:} N/A \quad \textbf{Past Medical History:} N/A \\
\textbf{Findings:} Diffusion restriction in the \textcolor{blue}{anterior circulation} territory. \\
\textbf{Impression:} Acute infarction. \\[-0.6em]
\noindent\rule{\linewidth}{0.4pt}
\vspace{0.2em}
\textbf{Generated report:} \\
\textbf{Clinical Presentation:} A 58-year-old male patient. \\
\textbf{NIHSS:} N/A \quad \textbf{Past Medical History:} N/A \\
\textbf{Findings:} Diffusion restriction in the \textcolor{blue}{anterior circulation} territory. \\
\textbf{Impression:} Acute infarction. \\[-0.6em]
\noindent\rule{\linewidth}{0.4pt}
\vspace{0.2em}
\textbf{Retrieved similar cases:} \\
$[1]$ \textcolor{blue}{Anterior circulation}, mild wedge-shaped (Score: \textcolor{blue}{0.95}) \\
$[2]$ \textcolor{blue}{Anterior circulation}, mild small striatocapsular (Score: \textcolor{blue}{0.93}) \\
$[3]$ \textcolor{blue}{Anterior circulation}, mild large vascular (Score: \textcolor{blue}{0.91}) \\
$[4]$ \textcolor{blue}{Anterior circulation}, mild large vascular (Score: \textcolor{blue}{0.91}) \\
$[5]$ \textcolor{blue}{Anterior circulation}, strong small (Score: \textcolor{blue}{0.90})
\end{minipage} \\
\bottomrule[1.5pt]
\end{tabular}
\end{table*}

Table~\ref{tab:qualitative} shows PIRTA generates reports closely aligning with expert references; higher retrieval similarity consistently corresponds to better factual accuracy, directly linking retrieval quality to report factuality.

\section{Discussion and Conclusion}

By reformulating report generation as an image-domain retrieval and text-domain augmentation task, PIRTA overcomes key limitations of direct image--text alignment for volumetric MRI, reducing optimization complexity in data-constrained settings.

\noindent\textbf{Clinical implications of retrieval-grounded factuality.} Ischemic-territory information directly informs reperfusion triage in acute stroke care, where misattribution of vascular territory can shift treatment decisions and patient outcomes. PIRTA delivers consistent territory accuracy across the internal cohort and two external cohorts (BRMH, ISLES) without recalibration, with the explicit cosine-similarity score acting as a per-case confidence cue that lets unreliable retrievals be flagged for radiologist review.

\noindent\textbf{Territory accuracy as a focused factuality surrogate.} N-gram metrics such as BLEU and ROUGE miss clinically important factual errors in radiology reports, motivating ischemic-territory accuracy as a clinically grounded surrogate. We emphasize that territory accuracy is a \emph{focused} factuality measure rather than a complete report-quality metric: it captures whether the most treatment-relevant attribute is preserved through generation but does not directly score lesion size, perfusion mismatch, hemorrhagic transformation, or report fluency. We adopt it as a high-signal, clinically actionable axis while leaving complementary metrics for future evaluation.

\noindent\textbf{Limitations and planned evaluation axes.} Factual accuracy degrades when rare or atypical infarct patterns are underrepresented in the retrieval database, yielding low cosine similarity and weak grounding. Beyond expanding the retrieval corpus, planned axes include (i) finer-grained ASPECTS-style scoring for per-region attribution, (ii) prospective clinical evaluation with radiologist-rated factuality on held-out reports, and (iii) hybrid multimodal retrieval that augments image embeddings with clinical metadata (NIHSS, presentation, PMH) to recover rare-pattern grounding.

In summary, PIRTA improves the factuality of 3D brain MRI report generation by grounding outputs in clinician-authored reports retrieved from similar cases, offering a scalable paradigm that achieves superior ischemic-territory accuracy without explicit cross-modal alignment.


\begin{credits}
\subsubsection{\ackname}
This work was supported by the National Research Foundation of Korea (NRF) grant funded by the Korea government (MSIT) (No. RS-2026-25479661) (K.S.C.); the SNUH Research Fund (No. 04-2024-0600; No. 04-2025-2060) (K.S.C.); the Korea Health Technology R\&D Project through the Korea Health Industry Development Institute (KHIDI) grant funded by the Ministry of Health\&Welfare (No. RS-2024-00439549) (K.S.C.); and the Bio\&Medical Technology Development Program of the National Research Foundation (NRF) funded by the Korean government (MSIT; RS-2025-02263045) (B.-H.K.).

\subsubsection{\discintname}
Byung-Hoon Kim serves as a director of EverEx, which is unrelated to the matter of this study. The other authors have no competing interests to declare that are relevant to the content of this article.
\end{credits}

\bibliographystyle{splncs04}
\bibliography{ref}

\end{document}